\documentclass[a4paper,fleqn]{cas-dc}

\usepackage[numbers]{natbib}
\usepackage{mathtools}
\usepackage{amsmath,amssymb}
\usepackage{amsfonts}
\usepackage{amsthm}
\usepackage{stmaryrd}
\usepackage{bm}
\usepackage{xcolor}
\usepackage{tikz}
\usepackage{caption}
\usepackage{subcaption}
\usepackage{pgfplots,filecontents}
\usepgfplotslibrary{groupplots}
\pgfplotsset{compat=newest}
\usepackage{graphicx}
\usepackage{float}
\usetikzlibrary{positioning}

\def\tsc#1{\csdef{#1}{\textsc{\lowercase{#1}}\xspace}}
\tsc{WGM}
\tsc{QE}
\tsc{EP}
\tsc{PMS}
\tsc{BEC}
\tsc{DE}
\newcommand{\mat}[1]{\mathbf{#1}}
\newcommand{\vect}[1]{\mathbf{#1}}
\newcommand{\vx}{\vect{x}}
\newcommand{\vm}{\bm{\mu}}
\newcommand{\vX}{\mat{X}}
\newcommand{\vS}{\mat{S}}
\newcommand{\R}{\mathbb{R}}
\newcommand{\G}{\mathcal{G}}
\newcommand{\norm}[1]{\left\lVert#1\right\rVert}
\DeclareMathOperator*{\argmax}{argmax}
\definecolor{mypurple}{rgb}{0.59, 0.44, 0.84}

\newtheorem{definition}{Definition}
\newtheorem{property}{Property}
\newtheorem*{remark}{Remark}

\begin{document}
\let\WriteBookmarks\relax
\def\floatpagepagefraction{1}
\def\textpagefraction{.001}
\shorttitle{Few-Shot Classification under Graph Structure Priors}
\shortauthors{M. Bontonou et~al.}

\title [mode = title]{Graph-LDA: Graph Structure Priors to Improve the Accuracy in Few-Shot Classification}

\author[1]{Myriam Bontonou}[orcid=0000-0002-0010-5457]
\cormark[1]
\ead{myriam.bontonou@imt-atlantique.fr}
\credit{Conceptualization, Methodology, Code, Experiments, Writing}
\address[1]{IMT Atlantique, Lab-STICC, UMR CNRS 6285, F-29238, France}

\author[1]{Nicolas Farrugia}[orcid=0000-0002-1159-3513]
\ead{nicolas.farrugia@imt-atlantique.fr}
\credit{Conceptualization, Writing}

\author[1]{Vincent Gripon}[orcid=0000-0002-4353-4542]
\ead{vincent.gripon@imt-atlantique.fr}
\credit{Conceptualization, Writing}

\cortext[cor1]{Corresponding author}

\begin{abstract}
It is very common to face classification problems where the number of available labeled samples is small compared to their dimension. These conditions are likely to cause underdetermined settings, with high risk of overfitting. To improve the generalization ability of trained classifiers, common solutions include using priors about the data distribution. Among many options, data structure priors, often represented through graphs, are increasingly popular in the field. In this paper, we introduce a generic model where observed class signals are supposed to be deteriorated with two sources of noise, one independent of the underlying graph structure and isotropic, and the other colored by a known graph operator. Under this model, we derive an optimal methodology to classify such signals. Interestingly, this methodology includes a single parameter, making it particularly suitable for cases where available data is scarce. Using various real datasets, we showcase the ability of the proposed model to be implemented in real world scenarios, resulting in increased generalization accuracy compared to popular alternatives.
\noindent
\end{abstract}

% \begin{highlights}
% \item Research highlights item 1
% \item Research highlights item 2
% \item Research highlights item 3
% \end{highlights}


\maketitle

\section{Introduction}

During the last decade, significant advances have been achieved in classification, notably thanks to the popularization of Deep Neural Networks (DNN)~\cite{krizhevsky2012imagenet}. Among other factors, the success of DNN can be explained by their ability to benefit from very large training datasets. Models that contain a large number of parameters are likely to reach better performance, provided that they are trained with sufficient data, as shown in the trends in the field of computer vision~\cite{kornblith2019better} and natural language processing~\cite{wolf2020transformers}. Problematically, in some application domains, the acquisition of data can be prohibitively costly, leading to cases where large models have to be trained using scarce datasets.

To answer this problem, a very active field of research called few-shot learning has seen numerous developments in the last few years~\cite{wang2020generalizing}. Contributions can be broadly categorized in three main directions: meta learning~\cite{finn2017model, DBLP:conf/iclr/AntoniouES19}, hallucination of new examples~\cite{DBLP:conf/cvpr/ZhangZK19} and embedding based solutions~\cite{wang2019simpleshot, mangla2020charting}, the latter typically reaching the peak performance. In all cases, the key idea is to exploit a large generic dataset to help in the resolution of the task at hand. Mainly, these solutions rely on the use of a feature extractor trained using the previously mentioned generic dataset. Consequently, data is not considered raw, but rather transformed into easily-separable representations. Then, simple classifiers (e.g. logistic regression, nearest class mean\dots) are trained on the obtained representations. Depending on the quality of the feature extractor, it becomes possible to achieve competitive accuracy even when using only a few training examples for each class of the considered problem~\cite{mangla2020charting}. When it is complicated to have access to a generic dataset, such simple classifiers can be used directly on raw data, or on hand-crafted features.

In order to boost the accuracy of a classifier using few training examples, the idea of exploiting external information has become increasingly popular~\cite{bronstein2021geometric}. In many cases, this external information takes the form of a graph connecting the data samples of the considered dataset. In some applications, another form of external information consists of a model describing the inner structure of each data sample. Such cases are often encountered in the fields of neuroimaging~\cite{zhang2021functional} or biology~\cite{spalevic2020hierarchical}. Such a structure can also be modeled by a graph, as it has been shown recently in the field of Graph Signal Processing (GSP)~\cite{shuman2013emerging,ortega2018graph}.

The main motivation of this work is to address the following question: how to efficiently exploit prior knowledge about the structure of a multivariate signal in the context of few-shot classification? Our main contribution is to introduce a simple model requiring a graph data structure prior, that can be entirely described using a single parameter and that is well suited to real world classification datasets, resulting in statistically significant improvements in accuracy.

More precisely, we consider a setting where each data sample is the result of the sum of three components: a) a fixed class centroid, b) an isotropic graph-agnostic noise and c) a graph-dependent noise. We motivate this setting by considering that, in many cases, the graph noise can model complex intrinsic sources of correlation in the observed data, and the isotropic noise accounts for external sources of noise, for example related to the acquisition process and measurement error. Under our model, the classification problem amounts to separating Gaussian distributions having the same covariance matrix. As such, an optimal solution consists in whitening the data (i.e. orthogonalizing the dimensions and normalizing the new dimensions) before separating them with a hyperplane. This popular approach is known under the name Linear Discriminant Analysis (LDA). 
If, in general, LDA is not efficient in the context of this work where too few accessible examples prevent to reliably estimate the covariance matrix of the classes, our model takes advantage of the graph data structure prior to whiten the samples, hence the name graph-LDA. Only a single parameter is required to significantly boost the accuracy in real world problems.

Our contributions are the following:
\begin{itemize}
    \item We introduce a model to classify multivariate signals under data structure priors in a few-shot context;
    \item We show that, under this model, the optimal solution is a particular case of LDA, which we call graph-LDA;
    \item We show that graph-LDA improves the performance on synthetic data and we validate the usability of the model in real world scenarios using computer vision and neuroimaging benchmarks. The code to reproduce the experiments is available at \url{https://github.com/mbonto/graph-LDA}.
\end{itemize}

The article is organized as follows. Related works are presented in section~\ref{sec:relatedwork}. The model taking into account data structure priors is formally stated in section~\ref{sec:modeling}. The graph-LDA solution is derived in section~\ref{sec:methods}. Results on synthetic data and real-world data are presented in section~\ref{sec:exp}. Section~\ref{sec:ccl} is a conclusion.

\section{Related work}
\label{sec:relatedwork}
In this work, we consider a classification problem in which we have access to a small number of samples for each class, each one being subject to two sources of noise: an isotropic one and a graph-colored one.

When the ratio between the number of training samples and the number of features is small, it is often advised to use simple classifiers (i.e. classifiers with few parameters to learn) with prior knowledge~\cite{vignac2020choice} in order to avoid overfitting. Thus, in a few-shot context, it is particularly relevant to take advantage of any external information about the data. In this work, we consider having access to such information taking the form of a graph connecting components of the considered data samples. 

It is really common to use graphs to model the dependencies between features, and in particular, to model noise~\cite{segarra2017network, segarra2016blind}. A large number of methods have been developed to classify signals structured by graphs~\cite{zhou2020graph}, going from linear graph filters followed by logistic regressions~\cite{wu2019simplifying} to DNNs adapted to exploit graph structures~\cite{DBLP:conf/nips/DefferrardBV16}. In particular, in the last few years, there has been many contributions in the field of Graph Signal Processing (GSP), a mathematical framework meant to generalize the notion of frequency analysis to graph structured data. 

Under the GSP framework, a graph signal is defined as follows. A graph $\G$ is composed of vertices $\mathcal{V}=\left\{1, \dots, d \right\}$. A graph signal is a vector $\vect{x}\in \mathbb{R}^\mathcal{V}$ whose each component is associated with a vertex of $\G$. The edges $\mathcal{E}$ of $\G$ are pairs of vertices $(i, j)$ associated with a weight. The adjacency matrix $\mat{A}\in \mathbb{R}^{\mathcal{V} \times \mathcal{V}}$ is defined using its elements $\mat{A}_{i, j}$ as the weight of the edge $(i, j)$. In many cases, the graph is undirected, and as such its adjacency matrix is symmetric. Such a graph can be used as a model of the relationships between the components of a signal. Sometimes, authors define the notion of Graph-Shift Operator (GSO) $\vS$ as any symmetric matrix that shares the same support as $\mat{A}$. In our work, we shall consider that $\vS$ is either the adjacency matrix of the graph or any GSO.

Based on $\vS$, it is common to use the so-called Graph Fourier Transform (GFT) of a signal $\vect{x}$ over the graph $\G$. We make use of the GFT in this work. It is defined as follows. First, let us point out that since $\vS$ is symmetric, one can use the spectral theorem to write $\vS = \mat{U}\mathbf{\Lambda}\mat{U}^\intercal$, where $\mathbf{\Lambda} \in \R^{\mathcal{V} \times \mathcal{V}}$ is a diagonal matrix containing the eigenvalues $\lambda_1, \dots , \lambda_d$ in ascending order and $\mat{U} = [\vect{u}_1, \dots , \vect{u}_d] \in \R^{\mathcal{V} \times \mathcal{V}}$ is an orthonormal matrix whose columns are the corresponding eigenvectors. The GFT of a graph signal $\vx$ is defined as $\hat{\vx} = \mat{U}^\intercal\vx$. 

In the broader context of few-shot learning, authors mostly rely on the use of transfer learning techniques~\cite{mangla2020charting, wang2019simpleshot, DBLP:conf/iclr/AntoniouES19}. In transfer learning, a feature extractor is trained on a generic dataset before being used to embed the data samples of the few-shot problem in a latent space where they are more easily classified. Most of these approaches do not consider any priors about data structure in the considered latent space. Very simple classifiers are typically used~\cite{wang2019simpleshot}, such as the Nearest Class Mean classifier (NCM) where each class is represented a centroid (the average of its training examples) and the class assigned to a new input is that of the nearest centroid.

Graph-LDA, the optimal solution for our model, is based on LDA. LDA is a simple classifier known to be optimal when classes follow Gaussian distributions with the same covariance matrix.
Using LDA, the challenge, in particular with few samples, is to estimate the covariance matrix $\bm{\Sigma}$. 
When few samples are available, a recommended estimator is a shrinkage estimator $\delta\vect{F} + (1 - \delta)\hat{\bm{\Sigma}}$, where $\hat{\bm{\Sigma}}$ is the empirical covariance matrix, $\vect{F}$ is a structure estimator and $\delta$ is the shrinkage coefficient. The sample covariance matrix is shrunk towards the structure estimator. When classes have a Gaussian distribution, the oracle approximating shrinkage estimator (OAS) is recommended~\cite{chen2010shrinkage} to define $\vect{F}$ and $\delta$. 
Contrary to LDA, our proposed solution does not need to estimate the whole covariance matrix since this one is characterized using a single parameter. In the Experiments section, we compare the accuracy of our graph-based solution with the one of LDA using OAS.

An example of application for our model is that of brain activity decoding from functional Magnetic Resonance Imaging (fMRI). Indeed, fMRI measurements are high dimensional (up to 200000 dimensions), and are intrinsically very noisy, as the signal of interest constitutes only a fraction of the measures~\cite{poldrack2011handbook}. However it is often possible to use other imaging techniques, as well as previous studies, to build a graph that models dependencies between brain areas~\cite{bassett2018nature}. Several prior studies have shown the potential of using transfer learning~\cite{mensch2021extracting,bontonou2020few} or GSP to analyze fMRI data~\cite{Huang2018,lioi2021gradients}, resulting in better classification of signals using subsets of the obtained features~\cite{Menoret_2017,pilavci2019spectral,brahim2020graph}.

It is worth mentioning that LDA is very commonly used in brain activity decoding. An early reference to LDA for Brain Computer Interfaces (BCI) based on Electroencephalography (EEG) can be found in~\cite{pfurtscheller1999eeg}. Since then, LDA has been widely used in a variety of EEG-BCI settings, as first reviewed in~\cite{lotte2007review}, and is still a method of reference in the last years~\cite{lotte2018review}. LDA has also been successfully applied to classify fMRI data, notably as a method of choice in the literature on the so called "Multi Voxel Pattern Analysis" (MVPA) approaches~\cite{haynes2005predicting,norman2006beyond}.

\section{Problem Statement}
\label{sec:modeling}
In machine learning, solving a classification problem consists in assigning labels to samples by learning from labeled training data. In this work, we are interested in a specific classification problem, in which the samples comply with the model presented thereafter.

Let us consider a graph $\G$ whose vertices are $\mathcal{V}$ and an associated GSO $\vS$. We suppose that each class $c$ is centered around a centroid vector $\vm_c\in \mathbb{R}^\mathcal{V}$. We additionally consider two model parameters $\alpha$ and $\beta$ and two random vectors indexed by $\mathcal{V}$ following multivariate normal distributions $\mat{N}_0\sim \mathcal{N}(\mat{0}, \mat{I})$ and $\mat{N}_1\sim \mathcal{N}(\mat{0}, \mat{I})$, $\mat{0}$ and $\mat{I}$ being respectively the null matrix and the identity matrix. A sample $\vect{x}$ in class $c$ is a vector drawn from the random vector:
\begin{equation}
\label{Eq:Model}
\vX_c = \vm_c + \alpha\vS\mat{N}_1 + \beta\mat{N}_0\;,
\end{equation}
Equivalently, the random vector characterizing the class $c$ follows a multivariate normal distribution with mean $\vm_c$ and covariance matrix $\bm{\Sigma}$, where $\bm{\Sigma} = \alpha^2\vS^2 + \beta^2\mat{I}$. There are two sources of variance, one is colored by the operator $\vS$ and set by $\alpha$: it amounts for an internal source of correlation within data samples. As such, the graph edges can be interpreted as a source of mixing between the connected components of each data sample. The second source of noise is set by $\beta$ and accounts for an inherent noise, such as typically related to the acquisition process.

In our model, the GSO $\vS$ is supposedly known. The class centroids $\vm_c$, and the parameters $\alpha$ and $\beta$ are unknown. For sake of simplicity, we derive in the next paragraphs an optimal methodology considering that samples are evenly distributed among all considered classes.

\section{Graph-LDA: improving few-shot classification accuracy under graph data structure priors}
\label{sec:methods}
As in our model the classes follow multivariate normal distributions sharing the same covariance matrix, an optimal classifier is the Linear Discriminant Analysis (LDA). LDA can be performed as soon as the covariance matrix of the classes is properly estimated.
In the following paragraphs, after introducing a few definitions, we recall why LDA is an optimal classifier and we adapt its methodology to our model. We call the resulting classifier graph-LDA.

\begin{definition}[Optimal classifier]
An optimal classifier $h^\star$ maximizes the expected accuracy.
\end{definition}

\begin{definition}[Expected accuracy]
Given $\vx\in\R^\mathcal{V}$ a sample, $y\in \llbracket 1, C\rrbracket$ its label and $\vX$, $Y$ their respective random vector and variable. Given a classifier
$h: \R^\mathcal{V} \to \llbracket 1, C\rrbracket$. 
The expected accuracy of the classifier $h$ is defined by $\mathbb{E}_{\vX, Y}[\text{acc}(h(\vX), Y)]$ where:
\begin{align}
    \text{acc}(h(\vx), y) = \left\{ 
                                \begin{array}{rcr}
                                1 & \text{if } h(\vx) = y  \\
                                0 & \text{otherwise}
                                \end{array}
                            \right.\;.
\end{align}
\end{definition}

In the following properties, we use the notion of discriminative function to recall that the LDA is an optimal classifier for our model.
\begin{definition}[Discriminative function]
The discriminative function associated with the class $c$ is denoted $g_c$. A class $c$ is assigned to a sample $\vx$ if $\forall c' \neq c,\, g_c(\vx) \geq g_{c'}(\vx)$.
\end{definition}
We then particularize the methodology of the LDA to our model and we decompose the optimal classifier into three simple steps. All proofs are given in Appendix~\ref{appendixA}.

\begin{property}
\label{prop:optimal_classifier}
An optimal classifier $h^\star$ is such that it associates a sample $\vx$ with the class $c$ reaching the highest posterior probability $p_{Y|\vX}(c|\vx)$.
\end{property}

\begin{property}
\label{prop:discriminative_functions}
Within our model, a classifier $h^\star$ reaching the highest posterior probability is characterized by $C$ discriminative functions: 
\begin{equation}
    g_c(\vx) = \vect{w}_c^\intercal\vx + w_{c0} \text{ with } 
    \left\{
        \begin{array}{ll}
            \vect{w}_c = \bm{\Sigma}^{-1}\vm_c, \\
            w_{c0} = -\frac{1}{2}\vm_c^\intercal\bm{\Sigma}^{-1} \vm_c.
        \end{array}
    \right.
\end{equation}
\end{property}

This property explicitly defines the discriminative functions of the LDA. In the proof of the following property, we first recall that LDA amounts to whitening the data samples (i.e. linearly transforming their features so that the covariance matrix of the features becomes the identity matrix) before applying a NCM. We then show that the data samples can be easily whitened by being projected into the graph Fourier space, as the covariance matrix becomes diagonal.

\begin{property}
\label{prop:sphering_NCM}
We call graph-LDA the optimal classifier $h^\star$ that amounts to whitening the data before applying a NCM. Namely, graph-LDA can be described in three steps:
\begin{enumerate}
    \item \textbf{Computing the GFT} associated with $\vS$ for all signals. Given a signal $\vx$, $\hat{\vx} = \mat{U}^\intercal\vx$;
    \item \textbf{Normalizing the transformed signals $\hat{\vx}$}.
    Given the diagonal matrix $\mat{D} = \mathbf{\Lambda}^2 + \left(\frac{\beta}{\alpha}\right)^2\mat{I}$,
    $\breve{\vx} = \mat{D}^{-\frac{1}{2}}\hat{\vx}$;
    \item \textbf{Classifying the normalized signals $\breve{\vx}$ with a NCM}.
\end{enumerate}
\end{property}

\begin{remark}
In practice, we call $\sigma$ the ratio $\frac{\beta}{\alpha}$. To apply graph-LDA on a classification problem, only this single parameter need to be tuned.
\end{remark}

\section{Experiments and Results}
\label{sec:exp}
In this section, we solve several few-shot classification problems using prior information on the relationships between the dimensions of the input signals. Our first objective is to evaluate to what extent the performance can be increased with our solution compared to other classifiers. To that purpose, we consider suitable cases on simulated signals whose distribution follows the model in Eq.~\ref{Eq:Model}. Our second objective is to compare the performance of graph-LDA with other classifiers on two real classification problems. In these problems, the signals do not explicitly follow the data generation model; the relationship have to be estimated from the available priors.

\begin{remark}
In our experiments, we compare graph-LDA with three other classifiers: NCM, logistic regression (LR) and LDA. We use the versions implemented on scikit-learn~\cite{scikit-learn}, with their default parameters at the exception of the LDA used with the 'lqsr' solver and the OAS estimator (recommended with few training samples and Gaussian distributions).
\end{remark}

Notice that we consider problems with few training samples in which a model may be useful. With more training examples, a LDA classifier trained from scratch is likely to correctly estimate the covariance matrix without needing to fit a model. Similarly, a classifier learning more parameters such as a logistic regression (LR) is more likely to overfit with few training samples.

\subsection{Simulations}
\begin{figure}[t]
\centering
\begin{subfigure}[b]{0.32\textwidth}
\includegraphics[width=\textwidth]{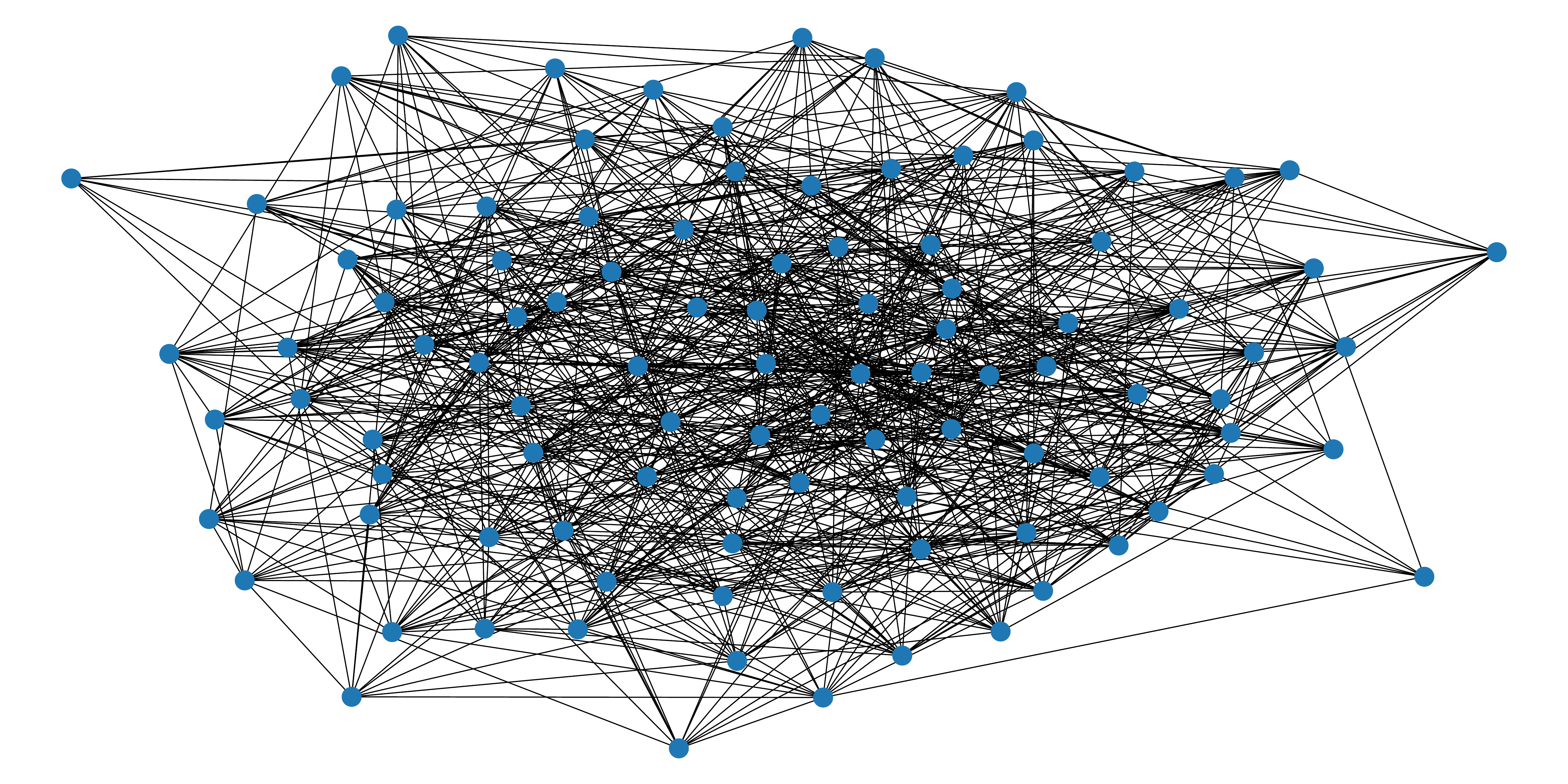}
\caption{Erdös-Renyi}
\end{subfigure}
\hfill
\begin{subfigure}[b]{0.32\textwidth}
\includegraphics[width=\textwidth]{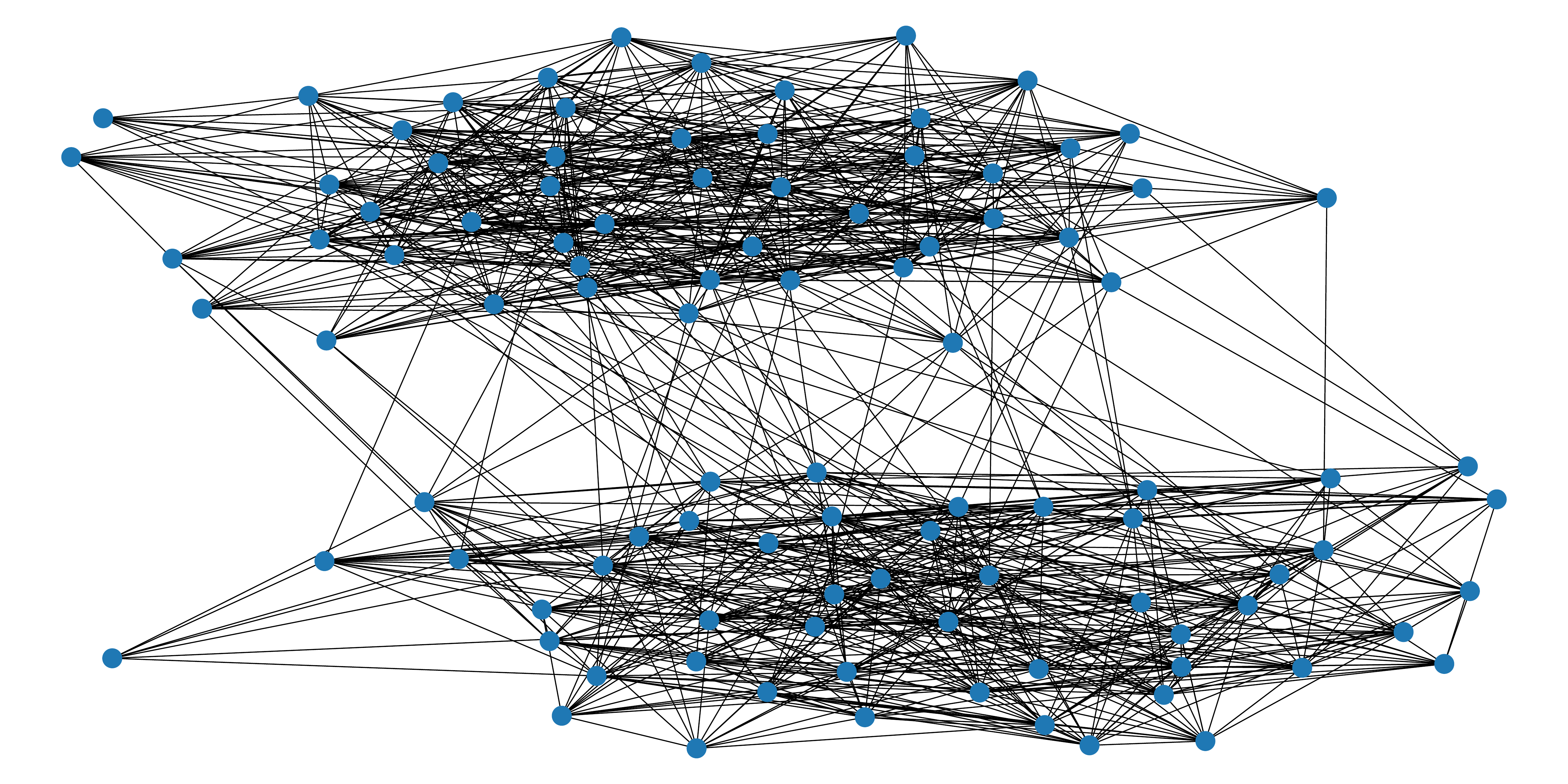}
\caption{Stochastic block model}
\end{subfigure}
\hfill
\begin{subfigure}[b]{0.32\textwidth}
\includegraphics[width=\textwidth]{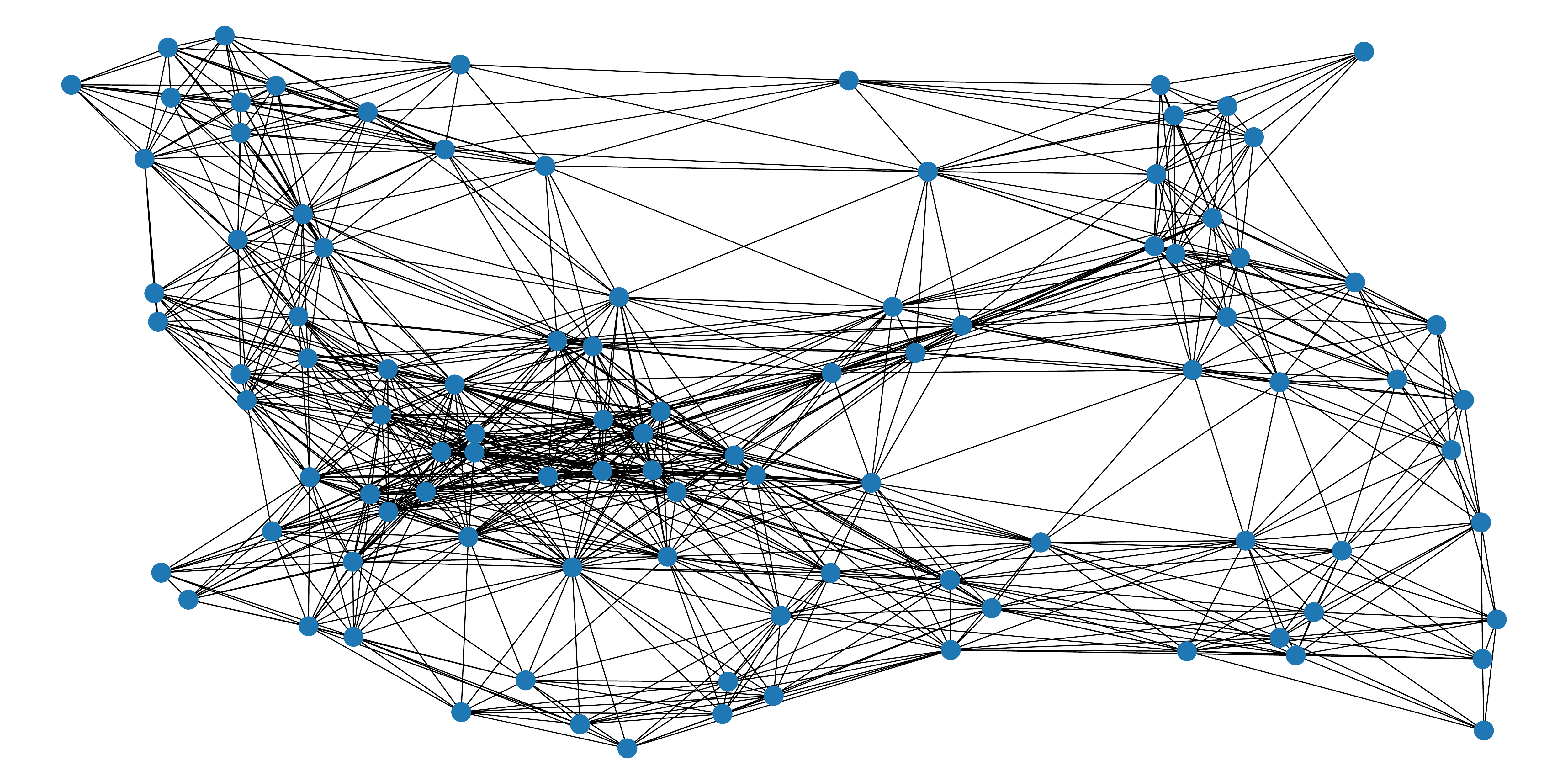}
\caption{Random geometric}
\end{subfigure}
\caption{Drawing of the three graphs used to simulate data samples.}
\label{fig:graphs}
\end{figure}

We propose several illustrative experiments on synthetic data in order to show to what extent the classification accuracy can be increased. For simplicity, we build datasets with two classes, and generate data samples following Equation~(\ref{Eq:Model}). 

\begin{figure*}[H]
\centering
\begin{tikzpicture}
\begin{groupplot}[group style={group size=3 by 1}]
\nextgroupplot[
        width=5.5cm,
        xlabel style={align=center},
	    xlabel=Number of training examples\\ \\(a) Erdös-Renyi,
	    ylabel=Accuracy,
	    ]
        \addplot [color=blue, line width=0.9pt, error bars/.cd, y dir = both, y explicit]
        table [x=shot, y=NCM, y error=NCM conf, col sep=comma] {./csv/varying_the_shots_erdo.csv};

        \addplot [color=blue, line width=2pt, dashed, error bars/.cd, y dir = both, y explicit]
        table [x=shot, y=NCM + opti, y error=NCM + opti conf, col sep=comma] {./csv/varying_the_shots_erdo.csv};

        \addplot [color=green, line width=0.9pt, error bars/.cd, y dir = both, y explicit]
        table [x=shot, y=LR, y error=LR conf, col sep=comma] {./csv/varying_the_shots_erdo.csv};

        \addplot [color=green, line width=1.5pt, dashed, error bars/.cd, y dir = both, y explicit]
        table [x=shot, y=LR + opti, y error=LR + opti conf, col sep=comma] {./csv/varying_the_shots_erdo.csv};

        \addplot [color=orange, line width=0.9pt, error bars/.cd, y dir = both, y explicit]
        table [x=shot, y=LDA, y error=LDA conf, col sep=comma] {./csv/varying_the_shots_erdo.csv};

        \addplot [color=orange, line width=0.9pt, dashed, error bars/.cd, y dir = both, y explicit]
        table [x=shot, y=LDA + opti, y error=LDA conf, col sep=comma] {./csv/varying_the_shots_erdo.csv};

\nextgroupplot[
        width=5.5cm,
        xlabel style={align=center},
	    xlabel=Number of training examples\\ \\(b) Stochastic block model,
	    ]
        \addplot [color=blue, line width=0.9pt, error bars/.cd, y dir = both, y explicit]
        table [x=shot, y=NCM, y error=NCM conf, col sep=comma] {./csv/varying_the_shots_stochastic_block_model.csv};

        \addplot [color=blue, line width=2pt, dashed, error bars/.cd, y dir = both, y explicit]
        table [x=shot, y=NCM + opti, y error=NCM + opti conf, col sep=comma] {./csv/varying_the_shots_stochastic_block_model.csv};

        \addplot [color=green, line width=0.9pt, error bars/.cd, y dir = both, y explicit]
        table [x=shot, y=LR, y error=LR conf, col sep=comma] {./csv/varying_the_shots_stochastic_block_model.csv};

        \addplot [color=green, line width=1.5pt, dashed, error bars/.cd, y dir = both, y explicit]
        table [x=shot, y=LR + opti, y error=LR + opti conf, col sep=comma] {./csv/varying_the_shots_stochastic_block_model.csv};

        \addplot [color=orange, line width=0.9pt, error bars/.cd, y dir = both, y explicit]
        table [x=shot, y=LDA, y error=LDA conf, col sep=comma] {./csv/varying_the_shots_stochastic_block_model.csv};

        \addplot [color=orange, line width=0.9pt, dashed, error bars/.cd, y dir = both, y explicit]
        table [x=shot, y=LDA + opti, y error=LDA + opti conf, col sep=comma] {./csv/varying_the_shots_stochastic_block_model.csv};

\nextgroupplot[
        width=5.5cm,
        xlabel style={align=center},
	    xlabel=Number of training examples\\ \\(c) Random geometric,
	    legend columns=-1,
	    legend style={at={(-2.25,1.1)}, anchor=south west, column sep=0.2cm},
	    ]
        \addplot [color=blue, line width=0.9pt, error bars/.cd, y dir = both, y explicit]
        table [x=shot, y=NCM, y error=NCM conf, col sep=comma] {./csv/varying_the_shots_random_geometric.csv};
        \addlegendentry{NCM}
        
        \addplot [forget plot, color=blue, line width=2pt, dashed, error bars/.cd, y dir = both, y explicit]
        table [x=shot, y=NCM + opti, y error=NCM + opti conf, col sep=comma] {./csv/varying_the_shots_random_geometric.csv};

        \addplot [color=green, line width=0.9pt, error bars/.cd, y dir = both, y explicit]
        table [x=shot, y=LR, y error=LR conf, col sep=comma] {./csv/varying_the_shots_random_geometric.csv};
        \addlegendentry{LR}
        
        \addplot [forget plot, color=green, line width=1.5pt, dashed, error bars/.cd, y dir = both, y explicit]
        table [x=shot, y=LR + opti, y error=LR + opti conf, col sep=comma] {./csv/varying_the_shots_random_geometric.csv};
        
        \addplot [color=orange, line width=0.9pt, error bars/.cd, y dir = both, y explicit]
        table [x=shot, y=LDA, y error=LDA conf, col sep=comma] {./csv/varying_the_shots_random_geometric.csv};
        \addlegendentry{LDA}
        
        \addplot [forget plot, color=orange, line width=0.9pt, dashed, error bars/.cd, y dir = both, y explicit]
        table [x=shot, y=LDA + opti, y error=LDA + opti conf, col sep=comma] {./csv/varying_the_shots_random_geometric.csv};
        
        \addlegendimage{color=gray, dashed, line width=0.9pt}
        \addlegendentry{With our preprocessing}
        \addlegendimage{color=gray, line width=0.9pt}
        \addlegendentry{Without}

\end{groupplot}
\end{tikzpicture}
\caption{Average accuracy and $95\%$ confidence interval over classification problems with $2$ classes generated from simulated datasets based on three different graphs. Each point is averaged over $100$ problems.}
\label{Fig:plots}
\end{figure*}
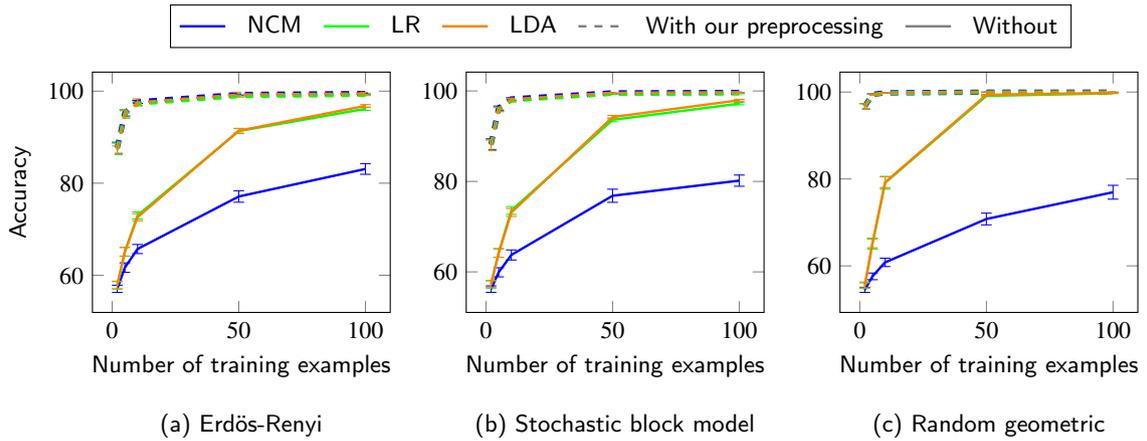

In this paragraph, we detail how we generate the class means and the data samples. The mean of the first class $\vm_1$ is the opposite of the mean of the second class $\vm_2$. Each coefficient of $\vm_1$ is obtained with a random variable following a uniform distribution on the interval $[-1, 1]$. As for the graph operator $\vS$, we experiment three types of random graphs: Erdös-Renyi, Stochastic Block Model (SBM), Random Geometric. Each graph has 100 nodes. The SBM has 2 clusters with 50 nodes in each. We define the edges probability such that each graph is expected to have the same number of edges (roughly 912 edges). More precisely, the probability of an edge between nodes $i$ and $j$ is $p_{ij} = 0.184$ for Erdös-Renyi. For the Random Geometric graphs, we set the radius to $0.274$. Finally, we use the probabilities $p_{ij} = 0.35 $ for SBM when $i$ and $j$ belong to the same cluster and $p_{ij} = 0.022$ otherwise. We assert that all graphs are connected. The matrix $\mat{S}$ is the adjacency matrix of these graphs. Figure~\ref{fig:graphs} shows a representation of these three graphs obtained using the NetworkX python package~\cite{schult2008exploring}. Finally, in Table~\ref{tab:simulated} and in Fig.~\ref{Fig:plots}, we set $\alpha = \beta = 1$, so that the ratio $\sigma$ between the two sources of noise is set to 1.

Using these simulated datasets, we first compare several standard classifiers (NCM, LR, LDA) using different preprocessing methods including: 1) our preprocessing method knowing the ratio $\sigma$ (ours), 2) dividing each sample by its norm (norm), 3) dividing each feature by its standard deviation computed on the training samples (std), or 4) computing the GFT of the signals and dividing each new feature by the standard deviation of this feature computed on the training samples (GFT + std). This last preprocessing method amounts to ours when there are enough examples to well estimate the standard deviations. In Tables~\ref{tab:erdo},~\ref{tab:random_geometric} and~\ref{tab:sbm}, we report the results over classification problems with two classes and 5 training samples per class. For each graph, the results are averaged over $100$ random classification problems with different randomly-generated class means.
For the three random graphs, we observe that, even if the performance of the NCM is not the best without preprocessing, adding our preprocessing step always enables to get one of the best classification performance. Our preprocessing step is more efficient than the other preprocessing steps, even when the ratio $\sigma$ is estimated from the data. It also helps the other classifiers to get better results.

In Fig.~\ref{Fig:plots}, we vary the number of training samples to explore the usefulness of our preprocessing step in different settings. 
In the three plots, we observe that it is always better to use our preprocessing step, regardless of the number of training samples or regardless of the classifier. We also see that our preprocessing step enables the NCM to get competitive results.

\begin{table*}[H]
\caption{Average accuracy and $95\%$ confidence interval over $100$ classification problems with $2$ classes and $5$ training examples per class generated from simulated datasets. Each table corresponds to a specific graph structure used to generate the data samples.}
\label{tab:simulated}
\begin{subtable}{\textwidth}
\centering
\subcaption{Erdös-Renyi graph.}
\vspace{-0.1cm}
\label{tab:erdo}
\begin{tabular}{|c|c|c|c|c|c|}
\hline
Classifier $\backslash$ Preprocessing & None & Ours & Spectral std & Std & Norm \\ 
\hline
NCM & $61.66 \pm 1.03$ & $\mathbf{95.26 \pm 0.69}$ & $90.43 \pm 0.68$ & $61.55 \pm 0.68$ & $62.53 \pm 0.68$ \\ 
LR & $65.07 \pm 0.96$ & $\mathbf{95.05 \pm 0.68}$ & $90.22 \pm 0.68$ & $63.98 \pm 0.68$ & $63.28 \pm 0.68$ \\ 
LDA & $65.12 \pm 0.94$ & $\mathbf{95.03 \pm 0.68}$ & $90.22 \pm 0.68$ & $64.1 \pm 0.68$ & $64.73 \pm 0.68$ \\ 
\hline
\end{tabular}
\end{subtable}
\begin{subtable}{\textwidth}
\centering
\vspace{0.3cm}
\subcaption{Random geometric graph.}
\vspace{-0.1cm}
\label{tab:random_geometric}
\begin{tabular}{|c|c|c|c|c|c|}
\hline
Classifier $\backslash$ Preprocessing & None & Ours & Spectral std & Std & Norm \\ 
\hline
NCM & $57.59 \pm 0.77$ & $\mathbf{99.47 \pm 0.13}$ & $98.22 \pm 0.14$ & $58.36 \pm 0.14$ & $58.65 \pm 0.14$ \\ 
LR & $65.03 \pm 1.15$ & $\mathbf{99.4 \pm 0.14}$ & $98.12 \pm 0.14$ & $63.89 \pm 0.14$ & $59.97 \pm 0.14$ \\ 
LDA & $65.27 \pm 1.1$ & $\mathbf{99.39 \pm 0.14}$ & $98.11 \pm 0.14$ & $64.34 \pm 0.14$ & $64.52 \pm 0.14$ \\ 
\hline
\end{tabular}
\end{subtable}
\begin{subtable}{\textwidth}
\centering
\vspace{0.3cm}
\subcaption{Stochastic block model graph.}
\vspace{-0.1cm}
\label{tab:sbm}
\begin{tabular}{|c|c|c|c|c|c|}
\hline
Classifier $\backslash$ Preprocessing & None & Ours & Spectral std & Std & Norm \\ 
\hline
NCM & $59.91 \pm 1.01$ & $\mathbf{96.21 \pm 0.45}$ & $92.03 \pm 0.48$ & $59.71 \pm 0.48$ & $60.49 \pm 0.48$ \\ 
LR & $64.19 \pm 0.95$ & $\mathbf{95.93 \pm 0.48}$ & $91.75 \pm 0.48$ & $63.01 \pm 0.48$ & $61.44 \pm 0.48$ \\ 
LDA & $64.16 \pm 0.94$ & $\mathbf{95.9 \pm 0.48}$ & $91.77 \pm 0.48$ & $63.15 \pm 0.48$ & $63.81 \pm 0.48$ \\ 
\hline
\end{tabular}
\end{subtable}
\end{table*}

\begin{figure*}[p]
\centering
\begin{tikzpicture}
  \node (img1)  {\includegraphics[width=0.41\textwidth]{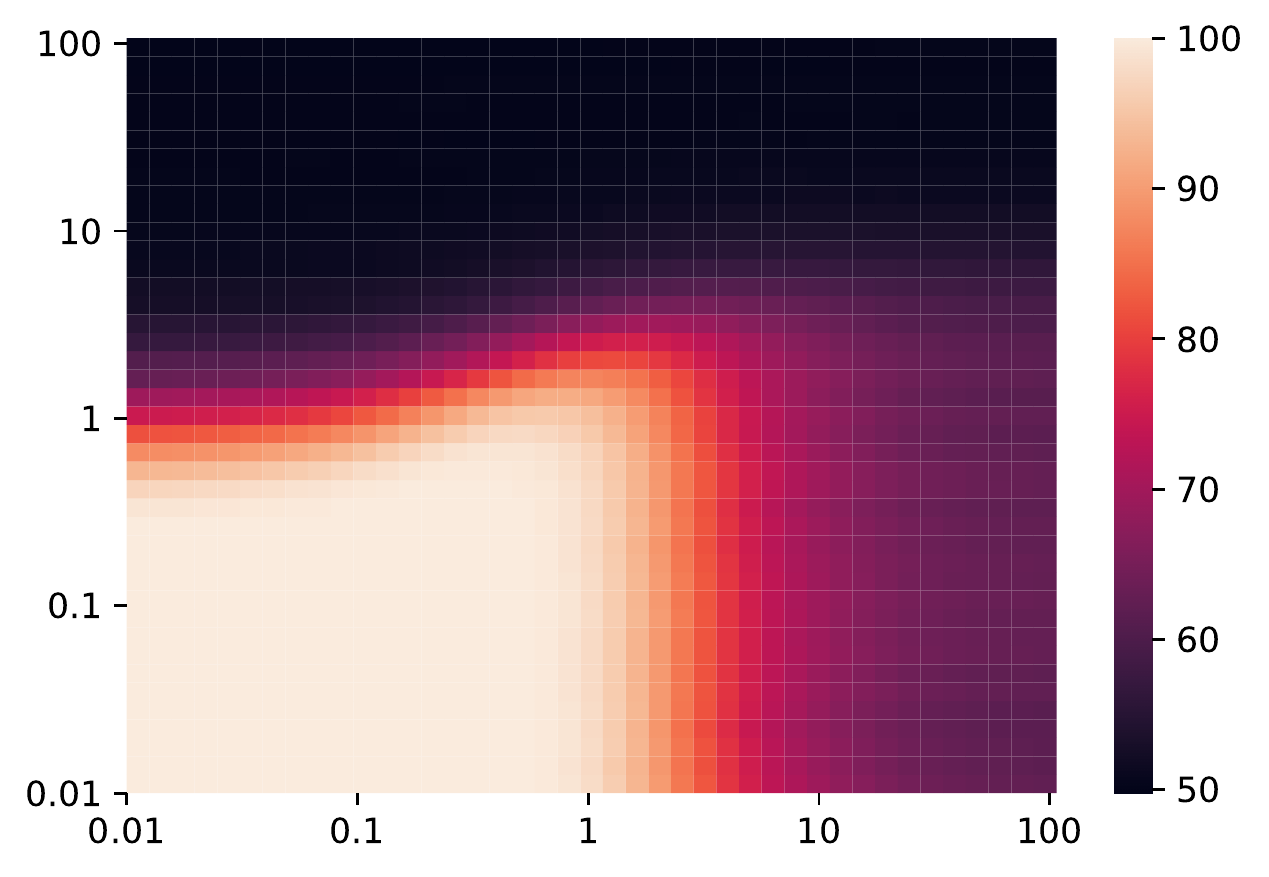}};
  \node[below=of img1, node distance=0cm, yshift=1cm] {Noise ratio $\tilde{\sigma}$ used to preprocess the data};
  \node[left=of img1, node distance=0cm, rotate=90, anchor=center, yshift=-0.7cm, text width=4.5cm, align=center] {Noise ratio $\sigma$ used to generate the data};
\end{tikzpicture}

\caption{Average accuracy obtained with graph-LDA (preprocessing + NCM) over $100$ classification problems with $2$ classes and $5$ training examples per class on the simulated dataset based on a Erdös-Renyi graph.}
\label{Fig:heatmaps}
\end{figure*}

In a last experiment, we explore the robustness of the performance to a wrong estimation of the ratio $\tilde{\sigma}$. In Fig.~\ref{Fig:heatmaps}, we report the average accuracy obtained on problems with $5$ training examples per class, generated using the simulated datasets based on the Erdös-Renyi graph with values of $\sigma$ varying between $0.01$ and $100$. When $\sigma$ is too large compared to the class means, the classification problem becomes too hard resulting in the fact that whatever the ratio, the accuracy is at the chance level. 
When $\tilde{\sigma}$ is equal to $\sigma \pm 1$, the accuracy is still really close to the optimal accuracy.
We obtained similar results for the two other random graphs.

\subsection{Decoding brain activity}
We now propose a first experiment on a real dataset: we train a classifier to decode brain activity from fMRI data. Decoding brain activity implies to recognize cognitive activities performed by people from brain signals, here measured in fMRI scanners. 

\begin{figure}
\centering
\includegraphics[width=0.45\textwidth]{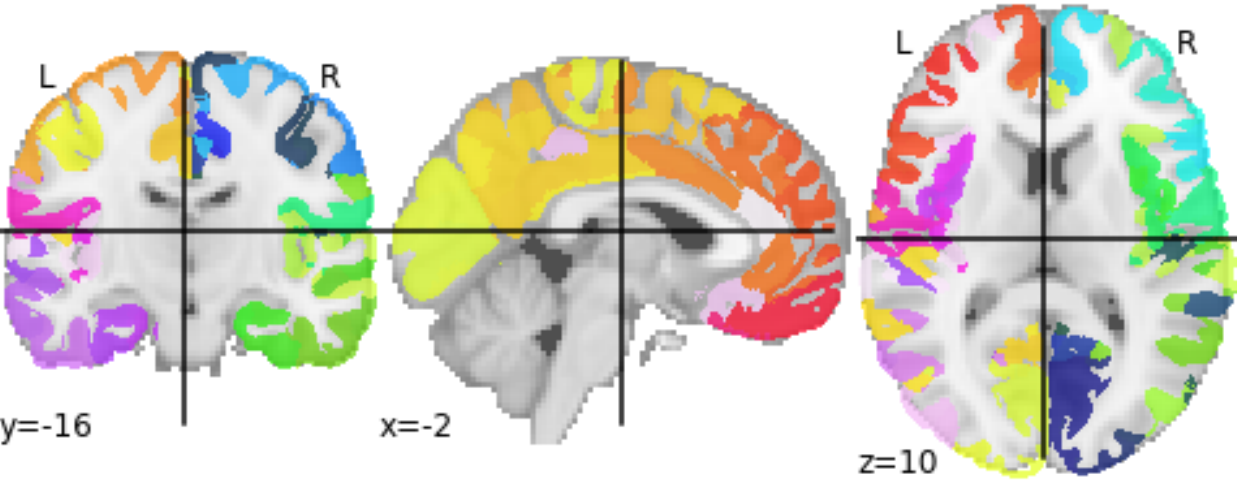}
\caption{Overview of the 360 regions of interest over which the brain activity is averaged.}
\label{Fig:roi}

\centering
\includegraphics[width=0.34\textwidth]{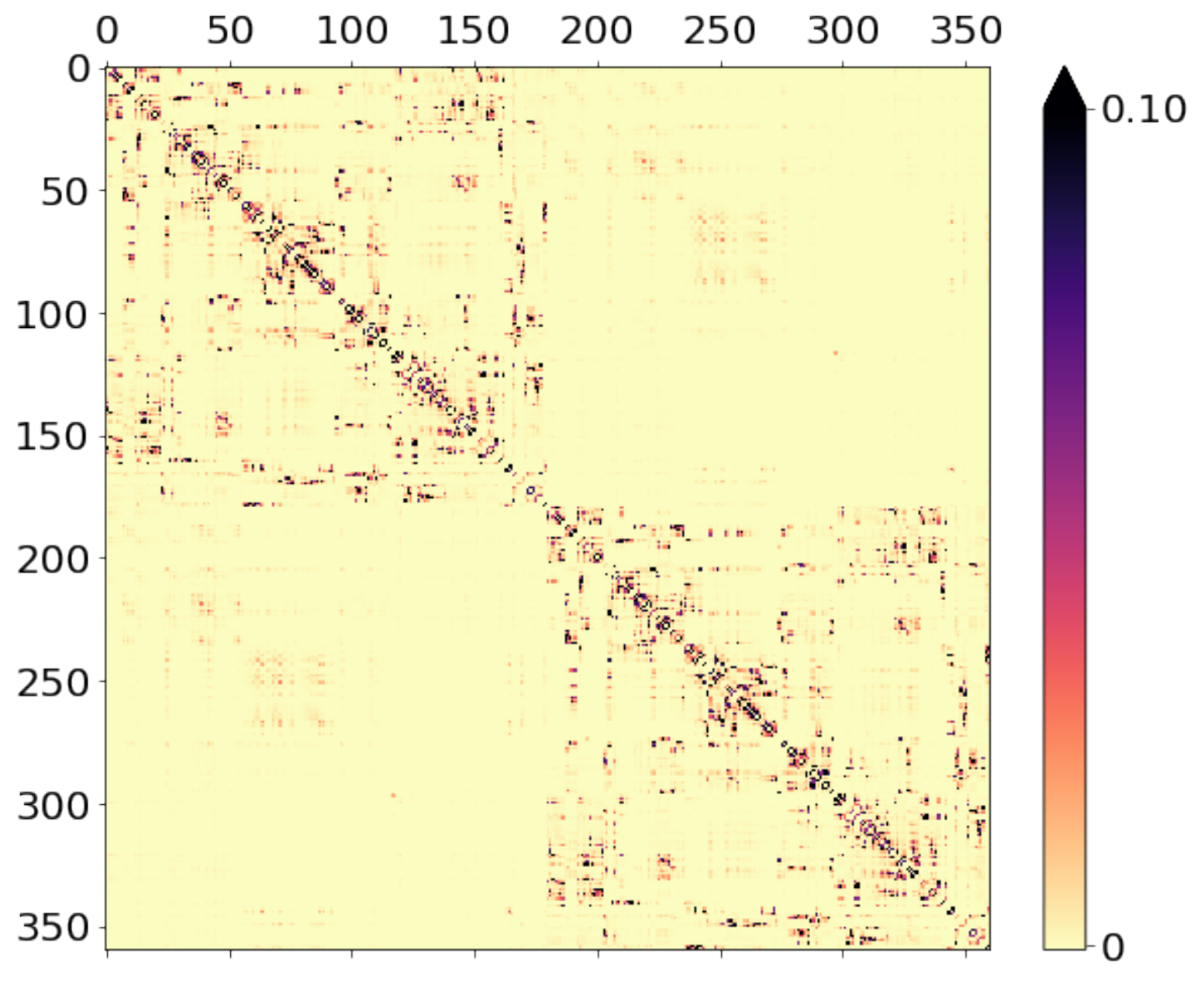}
\caption{Normalized adjacency matrix of brain connectivity used as an estimation of $\mat{S}$.}
\label{Fig:adjacency matrix}
\end{figure}

% Description of the data
Appendix~\ref{appendixB} contains a precise description of the dataset used in this article. In short, each data sample is an activation map derived from a general linear model, summarizing the changes in brain activity of a subject according to an experimental condition. The considered activation maps are spatially averaged into $360$ regions of interest (shown in Figure~\ref{Fig:roi}). For our model, we additionally use an estimation of the anatomical connectivity between those regions of interest. The connectivity is here considered the same across all subjects, and was measured in a previous study using white matter tractography~\cite{preti2019decoupling}. The corresponding graph is connected, and its adjacency matrix $\mathbf{A}$ is in $\R^{360\times 360}$. Given two brain regions $i$ and $j$, $\mathbf{A}[i, j]$ represents the strength of the anatomical connectivity between both regions of interest. 

% Experiment
From this dataset, we generate classification problems with $5$ classes, corresponding to a very common setting in few-shot learning. We compare the performance of a NCM with the ones of other classifiers, using or not our preprocessing step. Our preprocessing requires two parameters: the matrix $\mat{S}$ on which the noise is diffused and the ratio $\sigma$. 
We chose $\mathbf{S}$ such as $\mathbf{S} = \mathbf{D}^{-\frac{1}{2}}\mathbf{A}\mathbf{D}^{-\frac{1}{2}}$, where $\mathbf{D}$ is the degree matrix of $\mathbf{A}$. Intuitively, the diffused noise observed on a region $i$ is modeled as the noise of its neighboring regions weighted by the strength of their connection. 
As for the noise ratio, we tested several values and we chose $\tilde{\sigma} = 0.5$.

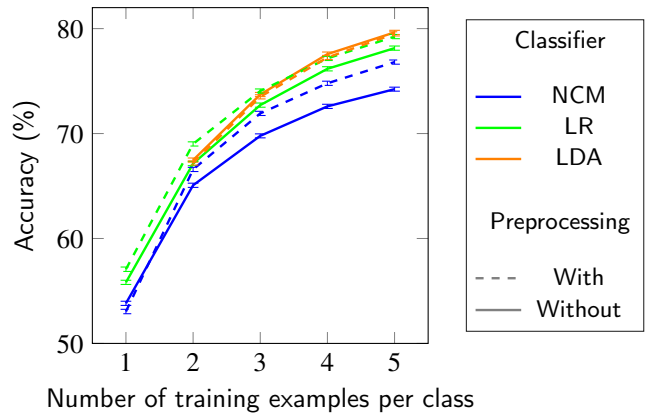
\begin{figure}
    \centering
    \pgfkeys{/pgf/number format/.cd,1000 sep={}}

\begin{tikzpicture}
\begin{groupplot}[group style={group size=2 by 1, horizontal sep=1cm, vertical sep=0cm}]

    \nextgroupplot[
    width=6cm, 
    height=6cm,
    ylabel near ticks, 
    xlabel near ticks,
	yticklabel style={font=\normalsize}, 
	xticklabel style={font=\normalsize}, %, rotate=45},
    xlabel style={font=\normalsize, align=center}, %, at={(0.55,-0.2)}}, 
    legend style={at={(0.95,0.05)}, anchor=south east}, 
    ylabel style={font=\normalsize, align=center}, 
    ylabel=Accuracy (\%), 
    xlabel=Number of training examples per class, 
    %xmode=log, 
    %log ticks with fixed point, 
    %scaled ticks=false, 
    xtick=data, 
    ymin=50, 
    ymax=82,
    xmin=0.5,
    xmax=5.5]

    \addplot[color=blue, line width=0.9pt, error bars/.cd, y dir = both, y explicit] 
    table [x=Shots, y=NCM, y error=NCM conf, col sep=comma] {csv/decoding_brain_activity.csv};
    
    \addplot[color=green, line width=0.9pt, error bars/.cd, y dir = both, y explicit] 
    table [x=Shots, y=LR, y error=LR conf, col sep=comma] {csv/decoding_brain_activity.csv};
    
    \addplot[color=orange, line width=0.9pt, error bars/.cd, y dir = both, y explicit] 
    table [x=Shots, y=LDA, y error=LDA conf, col sep=comma] {csv/decoding_brain_activity.csv};
    
    \addplot[color=blue, line width=0.9pt, dashed, error bars/.cd, y dir = both, y explicit] 
    table [x=Shots, y=NCM-prep, y error=NCM-prep conf, col sep=comma] {csv/decoding_brain_activity.csv};
    
    \addplot[color=green, line width=0.9pt, dashed, error bars/.cd, y dir = both, y explicit] 
    table [x=Shots, y=LR-prep, y error=LR-prep conf, col sep=comma] {csv/decoding_brain_activity.csv};
 
    \addplot[color=orange, line width=0.9pt, dashed, error bars/.cd, y dir = both, y explicit] 
    table [x=Shots, y=LDA-prep, y error=LDA-prep conf, col sep=comma] {csv/decoding_brain_activity.csv};

    \nextgroupplot[
    hide axis,
    width=3cm,
    height=3cm,
    ymin=0,
    ymax=0.4,
    xmin=0,
    xmax=10,
    legend style={font=\small, draw=white!15!black, at={(1.3,0.5)},anchor=east}
    ]
    
    \addlegendimage{empty legend}
    \addlegendentry{\hspace{-0.5cm}Classifier};
    
    \addlegendimage{empty legend}
    \addlegendentry{\tiny \color{white} *};
    
    \addlegendimage{color=blue, line width=0.9pt}
    \addlegendentry{NCM};
    
    \addlegendimage{color=green, line width=0.9pt}
    \addlegendentry{LR};
    
    \addlegendimage{color=orange, line width=0.9pt}
    \addlegendentry{LDA};
    
    \addlegendimage{empty legend}
    \addlegendentry{\color{white} *};
    
    \addlegendimage{empty legend}
    \addlegendentry{\hspace{-0.5cm}Preprocessing};
    
    \addlegendimage{empty legend}
    \addlegendentry{\tiny \color{white} *};
    
    \addlegendimage{color=gray, dashed, line width=0.9pt}
    \addlegendentry{With};
    
    \addlegendimage{color=gray, line width=0.9pt}
    \addlegendentry{Without};

\end{groupplot}
\end{tikzpicture}
    \caption{Average accuracy and $95\%$ confidence interval over $10000$ classification problems generated from a fMRI dataset. Each problem contains $5$ classes.}
    \label{Fig:neuro}
\end{figure}

% Describe the experiment
In Fig.~\ref{Fig:neuro}, we report the results we obtained with these parameters. Each accuracy is averaged over $10000$ classification problems, generated randomly from the set of available classes. Note that as the variance between the generated problems is more important, we average the accuracy over a larger number of problems than in the simulated experiments. For each problem, the accuracy is computed on $15$ unseen examples per class and a varying number of training examples. As it is not possible to estimate the covariance matrix with a single training example per class, we could not report the accuracy of the LDA in this case.
We observe that the preprocessing significantly boosts the accuracy of the NCM and the LR, and that LDA obtains peak accuracy when using three or more examples per class. Using our preprocessing, LR obtains better or equivalent accuracy than LDA. However, classification models based on NCM, while being more accurate using the proposed preprocessing, are globally less accurate than the other models. This might reflect the fact that the statistical distribution of examples in this dataset is suboptimally represented using the chosen graph structure, therefore the proposed whitening model is not well exploited by a NCM.

\subsection{Image classification}
In this section, we propose a second experiment on image classification problems generated from mini-ImageNet~\cite{vinyals2016matching}.

Mini-ImageNet is a dataset largely used in few-shot learning, composed of color images of size $64 \times 64$ separated into 100 classes. There are 600 images available per class. In standard few-shot learning~\cite{mangla2020charting}, the classes of the dataset are split into three sets: a base set, a validation set and a novel set. A DNN is trained on the base classes to learn representations of the samples that can be easily classified. The same DNN is evaluated on the validation classes, to assert that the way of representing the images is also adapted to a classification problem on new classes. The hyperparameters of the DNN are chosen according to the best performance on the validation set. Finally, new classification problems are generated from the novel classes. The DNN is used to generate representations of the examples. A new classifier is trained and tested on these representations. In this experiment, we propose to retrieve the representations of the data samples obtained at the output of a trained DNN and to directly classify the representations. To that end, we use the DNN trained in~\cite{mangla2020charting} obtaining state-of-the-art results.

From the novel classes of mini-ImageNet, we generate classification problems with $5$ classes.
To use graph-LDA (preprocessing + NCM) on this type of problems, we need to infer the matrix $\mathbf{S}$ and the ratio $\tilde{\sigma}$. Assuming that each sample $\vx$ follows the model described in Eq.~\ref{Eq:Model}, we related $\mathbf{S}$ to the empirical covariance matrix of the data samples, which is supposed to be equal to $\alpha^2\vS^2 + \beta^2\mat{I}$. We empirically estimated the covariance matrix of the data samples using the base set. Then, we chose to directly estimate the eigendecomposition of $\mathbf{S}$: $\vS$ and the empirical covariance matrix share the same eigenvectors and the eigenvalues of $\vS$ are approximated by the square roots of the eigenvalues of the covariance matrix. Then, using the validation set, it appeared that it was better to use a ratio $\tilde{\sigma} = 0.3$.

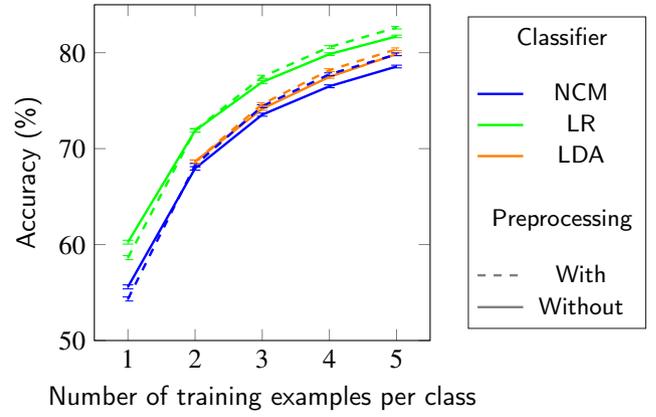
\begin{figure}
    \centering
    \pgfkeys{/pgf/number format/.cd,1000 sep={}}

\begin{tikzpicture}
\begin{groupplot}[group style={group size=2 by 1, horizontal sep=1cm, vertical sep=0cm}]

    \nextgroupplot[
    width=6cm, 
    height=6cm,
    ylabel near ticks, 
    xlabel near ticks,
	yticklabel style={font=\normalsize}, 
	xticklabel style={font=\normalsize}, %, rotate=45},
    xlabel style={font=\normalsize, align=center}, %, at={(0.55,-0.2)}}, 
    legend style={at={(0.95,0.05)}, anchor=south east}, 
    ylabel style={font=\normalsize, align=center}, 
    ylabel=Accuracy (\%), 
    xlabel=Number of training examples per class, 
    %xmode=log, 
    %log ticks with fixed point, 
    %scaled ticks=false, 
    xtick=data, 
    ymin=50, 
    ymax=85,
    xmin=0.5,
    xmax=5.5]

    \addplot[color=blue, line width=0.9pt, error bars/.cd, y dir = both, y explicit] 
    table [x=Shots, y=NCM, y error=NCM conf, col sep=comma] {csv/denoising_representations.csv};
    
    \addplot[color=green, line width=0.9pt, error bars/.cd, y dir = both, y explicit] 
    table [x=Shots, y=LR, y error=LR conf, col sep=comma] {csv/denoising_representations.csv};
    
    \addplot[color=orange, line width=0.9pt, error bars/.cd, y dir = both, y explicit] 
    table [x=Shots, y=LDA, y error=LDA conf, col sep=comma] {csv/denoising_representations.csv};
    
    \addplot[color=blue, line width=0.9pt, dashed, error bars/.cd, y dir = both, y explicit] 
    table [x=Shots, y=NCM-prep, y error=NCM-prep conf, col sep=comma] {csv/denoising_representations.csv};
    
    \addplot[color=green, line width=0.9pt, dashed, error bars/.cd, y dir = both, y explicit] 
    table [x=Shots, y=LR-prep, y error=LR-prep conf, col sep=comma] {csv/denoising_representations.csv};
 
    \addplot[color=orange, line width=0.9pt, dashed, error bars/.cd, y dir = both, y explicit] 
    table [x=Shots, y=LDA-prep, y error=LDA-prep conf, col sep=comma] {csv/denoising_representations.csv};

    \nextgroupplot[
    hide axis,
    width=3cm,
    height=3cm,
    ymin=0,
    ymax=0.4,
    xmin=0,
    xmax=10,
    legend style={font=\small, draw=white!15!black, at={(1.3,0.5)},anchor=east}
    ]
    
    \addlegendimage{empty legend}
    \addlegendentry{\hspace{-0.5cm}Classifier};
    
    \addlegendimage{empty legend}
    \addlegendentry{\tiny \color{white} *};
    
    \addlegendimage{color=blue, line width=0.9pt}
    \addlegendentry{NCM};
    
    \addlegendimage{color=green, line width=0.9pt}
    \addlegendentry{LR};
    
    \addlegendimage{color=orange, line width=0.9pt}
    \addlegendentry{LDA};
    
    \addlegendimage{empty legend}
    \addlegendentry{\color{white} *};
    
    \addlegendimage{empty legend}
    \addlegendentry{\hspace{-0.5cm}Preprocessing};
    
    \addlegendimage{empty legend}
    \addlegendentry{\tiny \color{white} *};
    
    \addlegendimage{color=gray, dashed, line width=0.9pt}
    \addlegendentry{With};
    
    \addlegendimage{color=gray, line width=0.9pt}
    \addlegendentry{Without};

\end{groupplot}
\end{tikzpicture}
    \caption{Average accuracy and $95\%$ confidence interval over $10000$ classification problems with $5$ classes and a various number of training examples per class generated from mini-ImageNet.}
    \label{Fig:ss}
\end{figure}

In Fig.~\ref{Fig:ss}, we report the accuracies obtained on the novel classes with different classifiers with or without our preprocessing method. The reported accuracies are averaged over $10000$ classification problems. For each problem, the accuracy is computed on $15$ unseen examples per class. We observe that after two training examples, the preprocessing slightly improves the accuracy of the LR and of the NCM. However, the preprocessed NCM (our graph-LDA) is still less accurate than the LR. That means that the LR succeeds to exploit information that is either not taken into account in our model or too complex to be handled by a NCM.

\section{Conclusion}
\label{sec:ccl}
In this paper, we considered a particular set of classification problems, with few training examples and a graph prior on the structure of each input signal. We introduced a solution considering a data generation model where the classes follow multivariate Gaussian distributions with a shared covariance matrix depending on the graph prior. According to this model, the optimal classifier, called graph-LDA, is a particular case of LDA in which only a single parameter has to be tuned. Graph-LDA can be decomposed into three simple steps: computing the GFT of the signals, normalizing the transformed signals and training a NCM on the normalized signals.

We showed that graph-LDA improves the accuracy on simulated datasets. We evaluated the relevance of the proposed method on real datasets, for which we made the hypothesis that the two first steps of graph-LDA could be a useful preprocessing for other classifiers. Graph-LDA stays a good solution when the graph prior on the structure of the input signals is strong.

In future work, we would like to investigate how to automatically estimate the single parameter of graph-LDA. It would also be interesting to define a protocol able to measure the adequacy of our model to a specific classification problem.

\subsection*{Declaration of competing interest}
The authors declare that they have no known competing financial interests or personal relationships that could have appeared to influence the work reported in this paper. This research did not receive any specific grant from funding agencies in the public, commercial, or not-for-profit sectors.

\appendix
\section{Proofs of the properties}
\label{appendixA}

\begin{proof}[Proof of property~\ref{prop:optimal_classifier}]
We look for a classifier $h$ maximizing $\mathbb{E}_{\vX, Y}[\text{acc}(h(\vX), Y)]$. We denote $f_{\vX}$ the probability density function of the random vector $\vX$, $p_Y$ the probability mass function of the random variable $Y$ and $f_{\vX, Y}$ the joint probability density function of the couple $(\vX, Y)$. By noticing that $\forall \vx$, $\forall y$, $f_{\vX, Y}(\vx, y) = f_{\vX}(\vx)p_{Y|\vX}(y|\vx)$, we have:
\begin{equation}
\begin{split}
    \mathbb{E}_{\vX, Y}[\text{acc}(h(\vX), Y)] & = & \int_x \sum_y f_{\vX, Y}(\vx, y)\text{acc}(h(\vx), y)dx \\
                                               & = & \int_x f_{\vX}(\vx) \sum_y p_{Y|\vX}(y|\vx)\text{acc}(h(\vx), y)dx \\
                                               & = & \int_x f_{\vX}(\vx) \mathbb{E}_{Y|\vX=\vx}[\text{acc}(h(\vX), Y)]dx \\
                                               & = & \mathbb{E}_{\vX}[\mathbb{E}_{Y|\vX=\vx}[\text{acc}(h(\vX), Y)]] \;.
\end{split}
\end{equation}
Thus, maximizing $\mathbb{E}_{Y|\vX=\vx}[\text{acc}(h(\vX), Y)]$ for each sample $\vx$ would be optimal. By noticing that $\text{acc}(h(\vx), y)=1$ if $y=h(\vx)$ and $\text{acc}(h(\vx), y)=0$ otherwise, we have:
\begin{equation}
\begin{split}
    \mathbb{E}_{Y|\vX=\vx}[\text{acc}(h(\vX), Y)] & = & \sum_y p_{Y|\vX}(y|\vx)\text{acc}(h(\vx), y)\\
                                               & = & p_{Y|\vX}(h(\vx)|\vx)\:.
\end{split}
\end{equation}
So, the optimal classifier $h^\star$ is the one attributing to each sample $\vx$ the class 
\begin{equation}
    h^\star(\vx)=\argmax_{c\in \llbracket 1, C\rrbracket}(p_{Y|\vX}(c|\vx))\;.
\end{equation}
\end{proof}

\begin{proof}[Proof of property~\ref{prop:discriminative_functions}]
We want to define the optimal classifier with discriminative functions. An optimal classifier attributes to a sample $\vx$ the class $c$ maximizing $p_{Y|\vX}(c|\vx)$.
According to the Bayes'theorem, 
\begin{equation}
p_{Y|\vX}(c|\vx) = \frac{p_{\vX|Y}(\vx|c)p_Y(c)}{p_{\vX}(\vx)}\;.
\end{equation}
As $\log$ is an increasing function, we can also choose the class $c$ as the one maximizing 
\begin{equation}
\log(p_{Y|\vX}(c|\vx)) = \log(p_{\vX|Y}(\vx|c)) + \text{cst}_1\;.
\end{equation}
Here, we call $\text{cst}_1 = \log(p_Y(c)) - \log{p_{\vX}(\vx)}$ because it has the same value for all $c$ (assuming that the classes are balanced).
As the samples belonging to class $c$ follow a multivariate normal distribution with $\bm{\Sigma}$ being invertible (real symmetric), we can write:
\begin{equation}
p_{\vX|Y}(\vx|c) = \frac{1}{(2\pi)^\frac{|\mathcal{V}|}{2}|\bm{\Sigma}|^\frac{1}{2}}
\exp{\left[ -\frac{1}{2}(\vx - \vm_c)^\intercal\bm{\Sigma}^{-1}(\vx - \vm_c)\right]}\;,
\end{equation}
where $|\mathcal{V}|$ is the number of vertices in $\mathcal{V}$. Thus,
\begin{equation}
\log(p_{\vX|Y}(\vx|c)) = -\frac{1}{2}(\vx - \vm_c)^\intercal\bm{\Sigma}^{-1} (\vx - \vm_c) + \text{cst}_2\;,    
\end{equation}
where $\text{cst}_2$ also has the same value for all $c$. By expanding this term and removing the constant values for all classes, we obtain $g_c(\vx) = \vect{w}_c^\intercal\vx + w_{c0}$ with $\vect{w}_c = \bm{\Sigma}^{-1}\vm_c$ and $w_{c0} = -\frac{1}{2}\vm_c^\intercal\bm{\Sigma}^{-1} \vm_c$. 
\end{proof}

\begin{proof}[Proof of property~\ref{prop:sphering_NCM}]
To simplify the discriminative functions, we can whiten the samples so that the covariance matrix of each class be equal to the identity matrix. Given a matrix $\mat{P}\in\R^{\mathcal{V}\times \mathcal{V}}$, as $\vX_c \sim \mathcal{N}(\vm_c, \bm{\Sigma})$, we know the distribution of the linearly transformed random vector $\mat{P}\vX_c$: $\mat{P}\vX_c \sim \mathcal{N}(\mat{P}\vm_c, \mat{P}\bm{\Sigma}\mat{P}^\intercal)$. Thus, 
if $\mat{P} = \bm{\Sigma}^{-\frac{1}{2}} = \alpha^{-1}\mat{D}^{-\frac{1}{2}}\mat{U}^\intercal$ (see related work for the spectral decomposition $\vS = \mat{U}\mathbf{\Lambda}\mat{U}^\intercal$), with $\mat{D}$ diagonal matrix such that $\mat{D}_{i,i} = \lambda_i^2 + \left(\frac{\beta}{\alpha}\right)^2$, the covariance matrix is diagonal. Indeed,
\begin{equation}
\begin{split}
\mat{P}\bm{\Sigma}\mat{P}^\intercal 
& = \alpha^2\mat{P}\vS^2\mat{P}^\intercal + \beta^2\mat{P}\mat{P}^\intercal \\
& = \alpha^2\alpha^{-2}\mat{D}^{-\frac{1}{2}}\mat{U}^\intercal \mat{U}\mathbf{\Lambda}^2\mat{U}^\intercal \mat{U}\mat{D}^{-\frac{1}{2}} + \beta^2\alpha^{-2} \mat{D}^{-\frac{1}{2}}\mat{U}^\intercal \mat{U} \mat{D}^{-\frac{1}{2}} \\
& = \mat{D}^{-1} \mathbf{\Lambda}^2 + \left(\frac{\beta}{\alpha}\right)^2 \mat{D}^{-1} \\
& = \mat{D}^{-1} (\mathbf{\Lambda}^2 + \left(\frac{\beta}{\alpha}\right)^2\mat{I}) \\
& = \mat{I}\;.
\end{split}
\end{equation}

Let us denote the projected samples $\breve{\vx} = \bm{\Sigma}^{-\frac{1}{2}}\vx$ and the projected class means $\breve{\vm}_c=\bm{\Sigma}^{-\frac{1}{2}}\vm_c$. The discriminative functions can be expressed as:
\begin{equation}
\begin{split}
g_c(\vx) & = \vect{w}_c^\intercal\vx + w_{c0} \\
& = (\vm_c^\intercal\bm{\Sigma}^{-\frac{1}{2}})(\bm{\Sigma}^{-\frac{1}{2}} \vx) -\frac{1}{2}(\vm_c^\intercal\bm{\Sigma}^{-\frac{1}{2}})(\bm{\Sigma}^{-\frac{1}{2}} \vm_c)\\
& = \breve{\vm}_c ^\intercal \breve{\vx} -\frac{1}{2} \breve{\vm}_c ^\intercal \breve{\vm}_c\;.
\end{split}
\end{equation}

Notice that the square distance between the mean of the class $c$ $\breve{\vm}_c$ and a sample $\breve{\vx}$ is:
\begin{equation}
\begin{split}
\norm{\breve{\vx} - \breve{\vm}_c}_2^2 
& = \breve{\vm}_c^\intercal\breve{\vm}_c - 2\breve{\vm}_c^\intercal\breve{\vx} + \breve{\vx}^\intercal\breve{\vx}\\
& = - 2 g_c(\vx) + \text{cst}_3\;,
\end{split}
\end{equation}
where $\text{cst}_3$ has the same value for all classes.
When we choose the class $c$ maximizing $g_c(\vx)$, we choose the class minimizing the square distance between $\breve{\vx}$ and the class means $\breve{\vm}_c$ ($\norm{\breve{\vx} - \breve{\vm}_c}_2^2$). This characterizes a NCM classifier (samples $\breve{\vx}$ mapped to the nearest class mean $\breve{\vm}_c$).

\begin{remark}
In practice, we only need to estimate one parameter to perform an optimal classification, which is the ratio $\sigma = \frac{\beta}{\alpha}$. 
Indeed, to perform the whitening step, one would need to estimate $\alpha$ and $\beta$. However, the NCM classifier is scale-invariant, meaning that closest centroids remain the same for any scaling of the samples.
\end{remark}

\end{proof}

\section{Neuroimaging dataset}
\label{appendixB}
The dataset we consider has been designed as a benchmark dataset for few-shot learning on brain activation maps. This dataset has been introduced in~\cite{bontonou2020few}, with the original data coming from the Individual Brain Charting (IBC) dataset releases 1 and 2~\cite{IBC2018, pinho2020individual}. IBC is a very large scale fMRI data collection in which twelve human subjects perform a wide variety of cognitive tasks in a fMRI scanner, spanned in many scanning sessions across several years. The fMRI data are extensively preprocessed to remove unwanted noise from the experiments (head movement, heart rate...) and are spatially normalized to the standard MNI-152 template representing a standard brain. All preprocessing steps are detailed in~\cite{IBC2018} and are based on automated, state of the art fMRI preprocessing techniques. 
For each participant and for each experiment, brain activation maps are estimated using a statistical model of the effect of a task on brain activity with respect to a baseline state. Each map is sampled as a 3D volume composed of voxels (volume elements). Voxels are averaged into 360 regions of interest defined by an atlas (Human Connectome Project multimodal parcellation~\cite{Glasser2016}). 
These regions of interest are connected by a network of white matter tracts whose density can be measured using tractography from diffusion weighted imaging. In this study, we consider an anatomical graph from a previous study obtained by averaging white matter connectivity measured in several people~\cite{preti2019decoupling}. The dataset has been used previously in~\cite{bontonou2020few} to apply few-shot learning methods on brain activation maps averaged per brain region.

\printcredits

\bibliographystyle{model1-num-names}

\bibliography{refs}

\bio{}
\textbf{Myriam Bontonou} received her M.Sc. degree in Biomedical Engineering with distinction from Imperial College London (United Kingdom) in 2018 and her engineering degree from Ecole Centrale de Nantes (France) in 2018 as well. She is currently a 3rd year PhD student in the Department of Mathematical and Electrical Engineering at IMT Atlantique (France). During 2020, she was an intern student at MILA, Université de Montréal (Canada). Her research interests include deep learning, machine learning and graph signal processing, with a particular interest for applications in neuroscience. 
\endbio

\bio{}
\textbf{Nicolas Farrugia} received his PhD in 2008. Nicolas is an assistant professor at IMT Atlantique in Brest, France, in the Mathematical and Electrical Engineering Department. He authored and coauthored more than 20 journal papers and 30 conference papers. He is co-leading the BRAIn project (Better Representations for Artificial Intelligence). His research interests include developing innovative methods to better understand the Brain using modern machine learning, deep learning as well as methods based on graphs such as graph signal processing. 
\endbio

\bio{}
\textbf{Vincent Gripon} is a permanent researcher with IMT Atlantique. He obtained his M.Sc. from École Normale Supérieure Paris-Saclay in 2008 and his PhD from IMT Atlantique in 2011. He spent one year as a visiting scientist at McGill University between 2011 and 2012 and he was an invited Professor at Mila and Université de Montréal from 2018 to 2019. His research mainly focuses on efficient implementation of artificial neural networks, graph signal processing, deep learning robustness and associative memories. He co-authored more than 70 papers in these domains.
\endbio

\end{document}